\documentstyle[eaclap,fleqn,leqno]{article}

\newcommand{\term}{{\tt term}}
\newcommand{\form}{{\tt form}}
\newcommand{\terms}{{\tt term}s}
\newcommand{\forms}{{\tt form}s}
\newcommand{\lingex}[1]{{\em #1}}

\newcommand{\qlfhat}{^{\wedge}}
\newcommand{\QLF}{\mbox{\it QLF}}

\newcommand{\substs}{\mbox{\it Subs}}
\newcommand{\varass}{\mbox{\it g}}
\newcommand{\Smerge}{\uplus}
\newcommand{\Replace}{\mbox{\it newexpr}}
\newcommand \qlfsub {\sqsupseteq}
\newcommand{\ix}[1]{\mbox{+#1}}
\newcommand{\ixof}[1]{\mbox{idx\_of(#1)}}
\newcommand{\reft}[1]{\mbox{rft(#1)}}

\newcounter{EEXAMPLE}

\newenvironment{AEXAMPLE}{\begin{list}{}
    {\topsep      0pt
     \partopsep   0pt
     \itemsep     .0ex
     
     \usecounter{EEXAMPLE}
     }\vskip-\lastskip}{\end{list}}
\newcommand{\ENUMA}[1]{\refstepcounter{equation}
\label{#1}\item[(\theequation)] \begin{AEXAMPLE}}
\newcommand{\ENDENUMA}{\end{AEXAMPLE}}

\newenvironment{EXAMPLE}{\begin{list}{}
    {\topsep      4pt
     \itemsep     .0ex
     \labelwidth  15pt
     \leftmargin  20pt
     
     }}{\end{list}}

\newcommand{\ENEW}[1]{\refstepcounter{equation}
\label{#1}\item[(\theequation)]}

\newcommand{\SREF}[1]{(\ref{#1})}

\newenvironment{quoteqlf}{\begin{small}\nopagebreak\begin{list}{}%
{\setlength{\leftmargin}{10pt}\setlength{\rightmargin}{0pt}}\item}%
{\end{list}\end{small}}

\title{Ellipsis and Quantification: A Substitutional Approach}

\author{Richard Crouch\\
        SRI International, Cambridge Computer Science Research Centre\\
        23 Millers Yard, Mill Lane,\\
        Cambridge, CB2 1RQ, UK
        \\{\tt rc@cam.sri.com}}

\begin{document}

\maketitle

\begin{abstract}
The paper describes a substitutional approach to ellipsis resolution giving
comparable results to \cite{DalShiPer:eahu}, but without the need for
order-sensitive interleaving of quantifier scoping and
ellipsis resolution. It is argued that the order-independence results
from viewing semantic interpretation as building  a description of a semantic
composition, instead of the more common view of interpretation as actually
performing the composition.
\end{abstract}

\section{Introduction}
\label{intro}

Dalrymple, Shieber and Pereira \shortcite{DalShiPer:eahu}  (henceforth, DSP)
give an equational treatment of ellipsis
via higher-order unification which, amongst other things, provides
an insightful  analysis of the interactions between ellipsis and
quantification.
But it suffers a number of drawbacks, especially when viewed from
a computational perspective.

First, the precise order in which quantifiers are scoped and ellipses
resolved determines the final interpretation of elliptical sentences.
It is hard to see
how DSP's analysis could be implemented within a system employing a
pipelined architecture that, say, separates quantifier scoping out from
other reference resolution operations---this would seem to preclude the
generation of some legitimate readings. Yet many systems, for good practical
reasons, employ this kind of architecture.

Second, without additional constraints, DSP slightly overgenerate readings
for sentences like
\begin{EXAMPLE}
\ENEW{ex1} John revised his paper before the teacher did, and so did Bill.
\end{EXAMPLE}
Kehler \shortcite{Kehler:adcafe} has convincingly argued  that this problem
arises because DSP do not distinguish between merely co-referential
and co-indexed (in his terminology, role-linked) expressions.

Third, though perhaps less importantly, higher-order unification going beyond
second-order matching is required for resolving ellipses involving
quantification.  This increases the computational complexity of the
ellipsis resolution task.

This paper presents a treatment of ellipsis which avoids these difficulties,
while having essentially the same coverage as DSP.
The treatment is easily implementable,
and forms the basis of the ellipsis resolution component currently
used within the Core Language Engine \cite{AlsCarCro:clare}.

Ellipsis interpretations are represented as simple sets of substitutions on
semantic representations of the antecedent.  The substitutions can be
built up in an order-independent way (i.e. before, after or during scoping),
and without recourse to higher-order unification.
The treatment is similar to the discourse copying analysis
of \cite{Kehler:adcafe}, and to the substitutional treatment
suggested by Kamp within Discourse Representation Theory, described
in \cite{GawPet:aaqiss}.  However, we extend the
notion of strict and sloppy identity to deal with more than just
pronouns.  In doing so, we readily deal with phenomena like scope
parallelism.

While the treatment of ellipsis is hopefully of some value in its own
right, a more general conclusion can be drawn concerning the requirements
for a {\em computational} theory of semantics.  Briefly, the standard
view within formal semantics, which DSP inherit, identifies
semantic interpretation with composition: interpretation is the process
of taking the meanings of various constituents and composing them together
to form the meaning of the whole.  This makes semantic interpretation
a highly order-dependent affair; e.g. the order in which a functor is
composed with its arguments can substantially affect the resulting meaning.
This is reflected in the order-sensitive interleaving of scope and
ellipsis resolution in DSP's account.  In addition, composition is
only sensitive to the {\em meanings} of its components.  Typically
there is a many-one mapping from compositions onto meanings. So, for
example, whether two terms with identical meanings are merely co-referential
or are co-indexed is the kind of information that may get lost: the
difference amounts to two ways of composing the same meaning.

The alternative proposed here is to view semantic interpretation as
a process of building a (possibly partial) {\em description} of the
intended semantic composition; i.e. (partial) descriptions of
what the meanings
of various constituents are,  and how they should be composed
together.\footnote{This is similar to Nerbonne's \shortcite{Nerbonne:cs}
constraint-based semantics, except that he
builds descriptions of logical forms, not semantic compositions.}
While the order in which composition operations
are performed can radically affect the outcome, the order in which
descriptions are built up is unimportant.  In the case of ellipsis,
this extra layer of descriptive indirection permits an equational
treatment of ellipsis that (i) is order-independent, (ii) can take account
compositional distinctions that do not result in meaning differences, and
also (iii) does not require the use of higher-order unification for dealing
with quantifiers.

The paper is organised as follows.  Section~\ref{substs} describes
the substitutional treatment of ellipsis by way of a few examples presented
in a simplified version of Quasi Logical Form (QLF)
\cite{AlsCro:msi,AlsCarCro:clare}.  Section~\ref{semantics} gives the
semantics for the notation, and argues that QLF is best understood as
providing descriptions of semantic compositions.  Section~\ref{parallelism}
raises some open questions concerning the determination of parallelism
between ellipsis and antecedent, and other issues.  Section~\ref{conc}
concludes.

\section{Ellipsis Substitutions}
\label{substs}

This section illustrates the substitutional treatment of ellipsis through
a small number of examples.  For presentation purposes we only
sketch the intended semantics of the simplified QLF notation used,
and a more detailed discussion is deferred until section~\ref{semantics}.

\subsection{Simple VP Ellipsis}
A simple, uninteresting example to fix some notation:
\begin{EXAMPLE}
\ENEW{ex2} John slept. So did Mary
\end{EXAMPLE}
We represent the first sentence, ignoring tense, as a (resolved) QLF
\begin{EXAMPLE}
\ENEW{Ante}
\begin{tabbing}
$[\ix{j}]:\mbox{sleep}(\;\term(\ix{j},\;\exists,\;$\=$
                               \lambda y.\mbox{name}(y,\mbox{`John'}),$\\
                              \>$\lambda y.y=\mbox{j\_smith}))$
\end{tabbing}
\end{EXAMPLE}
The noun phrase \lingex{John} gives rise to an existentially quantified
term, uniquely identified by the index $\ix{j}$.  The $\term$ expression
has four arguments: an index, a determiner/quantifier, an explicit restriction,
and an additional contextually derived restriction.  In this case, the
quantifier ranges over objects that are named `John' and are further
restricted to be identical to some (contextually salient) individual,
denoted by j\_smith. Prior to reference resolution, the contextual restriction
on the term would be an uninstantiated meta-variable; resolution consists
of instantiating meta-variables to contextually appropriate values.
The scope of the term is indicated by the scope node $[\ix{j}]:$
prefixing the formula $\mbox{sleep}(\;\term(\ix{j},\ldots))$.  Again,
prior to resolution this scope node would be an uninstantiated
meta-variable.

A generalized quantifier representation equivalent to the above is
\begin{EXAMPLE}
\ENEW{ex3}
\begin{tabbing}
$\exists($\=$
     \lambda y.(\mbox{name}(y,\mbox{`John'})\wedge y=\mbox{j\_smith}),$\\
  \>$\lambda x. \mbox{sleep}(x))$
\end{tabbing}
\end{EXAMPLE}
The index in the scope node means that to semantically
evaluate the QLF, you get hold of the quantifier, restriction and
contextual restriction of the corresponding term.  This forms a (generalized)
quantifier expression, whose body is obtained by discharging all occurrences
of the term and it index to a variable, and abstracting over the variable.
Terms and indices not dischargeable in this manner lead to uninterpretable
QLFs \cite{AlsCro:msi}.

We represent the elliptical sentence, again abbreviated, as a (partially
resolved) QLF:
\begin{EXAMPLE}
\ENEW{ex4}
\begin{tabbing}
$?P(\;\term(\ix{m},\;\exists,\;$\=$
       \lambda y.\mbox{name}(y,\mbox{`Mary'}),$\\
    \>$\lambda y.y=\mbox{m\_jones}))$
\end{tabbing}
\end{EXAMPLE}
$?P$ is an unresolved meta-variable.
To resolve the ellipsis, it
needs to be instantiated to some contextually salient predicate.

Along similar lines to DSP, we can set up an equation to determine
possible values for $?P$\footnote{Terms shown abbreviated, i.e.
$\term(\ix{j},\ldots)$ instead of
$\term(\ix{j},\;\exists,\;\lambda y.\mbox{name}(y,\mbox{`John'}),\;
\lambda y.y=\mbox{\tt j\_smith})$.}:
\begin{EXAMPLE}
\ENEW{ex5}
$?P(\term(\ix{j},\ldots)) =  \mbox{[\ix{j}]:sleep}(\term(\ix{j},\ldots))$
\end{EXAMPLE}
That is, we are looking for a predicate that when applied to the subject
term of the ellipsis antecedent returns the antecedent.  The interpretation
of the ellipsis is then given by applying this predicate to the subject
of the ellipsis.

The equation \SREF{ex5} is solved by setting $?P$ to something that takes a
term $T$ as an argument and substitutes $T$ for $\term(\ix{j},\ldots)$
and the index of $T$ for $\ix{j}$ throughout the ellipsis antecedent
(the RHS of \SREF{ex5}):
\begin{EXAMPLE}
\ENEW{ex6}
\begin{tabbing}
$?P =  T\qlfhat($\=$\mbox{[\ix{j}]:sleep}(\term(\ix{j},\ldots))$\\
                \>$\mid\{\term(\ix{j},\ldots)/T, \ix{j}/\ixof{T}\})$
\end{tabbing}
\end{EXAMPLE}
Here $T\qlfhat(\ldots)$ is a form of abstraction; for now it will do no
harm view it as a form of $\lambda$-abstraction, though this is not strictly
accurate. The substitutions are represented using the notation
`\mbox{$\mid\{old/new,\ldots\}$}'.

Applying this value for $?P$ in the ellipsis \SREF{ex4}, we get
\begin{EXAMPLE}
\ENEW{ex7}
$\mbox{[\ix{j}]:sleep}(\term(\ix{j},\ldots))$\\
$\mid\{\term(\ix{j},\ldots)/\term(\ix{m},\ldots), \ix{j}/\ix{m}\}$
\end{EXAMPLE}
Ellipsis resolution thus amounts
to selecting an antecedent and determining a set of substitutions
to apply to it.  For reasons that will be explained shortly, it is
important that resolution does not actually carry out the application
of the substitutions.  However, were we to do this in this particular case,
where the  antecedent \SREF{Ante} is fully resolved, we would successfully
capture the intended interpretation of the ellipsis, namely:
\begin{EXAMPLE}
\ENEW{ex8}
\begin{tabbing}
$[\ix{m}]:\mbox{sleep}(\;\term(\ix{m},\;\exists,\;$\=$
                               \lambda y.\mbox{name}(y,\mbox{`Mary'}),$\\
                              \>$\lambda y.y=\mbox{m\_jones}))$
\end{tabbing}
\end{EXAMPLE}
Note that the substitutions are
not applied in the conventional order; viz. first replace $\ix{j}$ by $\ix{m}$
throughout \SREF{Ante} and then replace $\term(\ix{j},\ldots)$ by
$\term(\ix{m},\ldots)$.  The first substitution would ensure that there
was no $\term(\ix{j},\ldots)$ for the second substitution to replace.
The order in which substitutions apply instead depends on the order in which
the expressions occur when making a top down pass through \SREF{Ante}, such
as one would do when applying semantic evaluation rules to the formula.

Note also that the term index substitution applies to the scope node,
so that $[\ix{j}]:$ is replaced by $[\ix{m}]:$.
This ensures that the term for Mary in the ellipsis gets a parallel scope to
the term for John in the antecedent. Scope parallelism may not be
significant where proper names are concerned, but is important when
it comes to more obviously quantificational terms (section~\ref{sc-para}).

\subsection{Evaluative Substitutions}

The meaning
of an ellipsis is composed in essentially the same way, and from the same
components, as the meaning of its antecedent.  However, some changes need to
be made in order to accommodate new material introduced by the ellipsis.
The substitutions specify what these changes are.  In the example
discussed above, the meaning of the ellipsis is built up in the same
way as for the antecedent, except that whenever you encounter a term
corresponding to `John' or something dependent/co-indexed with it, you
it is treated as though it were the term for `Mary' or
dependent/co-indexed with it.

This means that the substitutions act as directives controlling the way
in which QLF expressions within their scope are evaluated. They are
not syntactic operations on QLF expressions --- they are part of
the QLF object language.

The reason that substitutions are not `applied' immediately upon
ellipsis resolution is as follows.  At the time of deciding on the ellipsis
substitutions, the precise composition of the antecedent may not yet have
been determined.  (For instance the scopes of quantifiers or the contextual
restrictions on pronouns in the antecedent may not have been resolved;
this will correspond to the presence of uninstantiated meta-variables
in the antecedent QLF.)  The ellipsis should follow, modulo the substitutions,
the same composition as the antecedent, {\em whatever that composition
is eventually determined to be}.  It makes no sense to apply the
substitutions before the antecedent is fully resolved, though it does make
sense to decide what the appropriate substitutions should be.

In practical terms what this amounts to is exploiting re-entrancy in QLFs.
The elliptical QLF will contain a predicate formed from the antecedent
QLF plus substitutions.  Any uninstantiated meta-variables in the antecedent
are thus re-entrant in the ellipsis. Consequently, any further resolutions
to the antecedent are automatically imposed on the ellipsis.  This would not
be the case if the substitutions were treated as syntactic operations on
QLF to be applied immediately: some re-entrant meta-variables would
be substituted out of the ellipsis, and those remaining would not be subject
to the substitutions (which would have already been applied) when they were
eventually instantiated.

\subsection{Scope Parallelism}
\label{sc-para}

It was noted above that substitutions on term indices in scope nodes
ensures scope parallelism.  This is now illustrated with a more
interesting example (adapted from Hirshb\"uhler as cited by DSP).
\begin{EXAMPLE}
\ENEW{flag}
A Canadian flag hung in front of every house, and an American flag did too.
\end{EXAMPLE}
The antecedent has two possible scopings: a single Canadian flag in front
of all the houses, or each house with its own flag.  Whichever
scoping is given to the antecedent, a parallel scoping should be
given to the ellipsis.

A simplified QLF for \SREF{flag} is
\begin{EXAMPLE}
\ENEW{flag-qlf}
\begin{tabbing}
$?S1$: and(\=$?S2$: hang(\=$\term(\ix{c},\exists,\ldots),$\\
          \>           \>$\term(\ix{h},\forall,\ldots)),$\\
          \>$?P(\term(\ix{a},\exists,\ldots)))$
\end{tabbing}
\end{EXAMPLE}
where the indices $\ix{c}$, $\ix{a}$ and $\ix{h}$ are mnemonic for
Canadian flag, American flag and house.  Taking the first conjunct as the
antecedent, we can set up an equation
\begin{EXAMPLE}
\ENEW{flag-eq}
$?S2$:hang($\term(\ix{c}\ldots),\term(\ix{h}\ldots))$\\
= $?P(\term(\ix{c}\ldots))$
\end{EXAMPLE}
the solution to which is\footnote{In the next section we place some extra
constraints on possible solutions, but these aren't strictly relevant here.}
\begin{EXAMPLE}
\ENEW{flag-soln}
\begin{tabbing}
$?P = T\qlfhat$(\=$?S2$:hang($\term(\ix{c}\ldots),\term(\ix{h}\ldots))$\\
                \>$\mid\{\term(\ix{c}\ldots)/T, \ix{c}/\ixof{T}\}$
\end{tabbing}
\end{EXAMPLE}
This make the elliptical conjunct equivalent to
\begin{EXAMPLE}
\ENEW{flag-soln1}
$?S2$:hang($\term(\ix{c}\ldots),\term(\ix{h}\ldots))$\\
$\mid\{\term(\ix{c}\ldots)/\term(\ix{a},\ldots), \ix{c}/\ix{a}\}$
\end{EXAMPLE}

The scope node, $?S2$ can be resolved to $[\ix{h},\ix{c}]$
(`every house' takes wide scope), or $[\ix{c},\ix{h}]$ (`a Canadian flag'
takes wide scope).  Whichever resolution is made, the substitution of
$\ix{a}$ for $\ix{c}$ ensures parallel scoping in the ellipsis for
`an American flag'.  Cashing out the substitutions for the first case, we
have
\begin{EXAMPLE}
\ENEW{flag-qlf1}
\begin{tabbing}
$[]$:and(\=$[\ix{h},\ix{c}]$:hang(\=$\term(\ix{c},\exists,\ldots),$\\
         \>                       \>$\term(\ix{h},\forall,\ldots)),$\\
         \>$[\ix{h},\ix{a}]$:hang(\>$\term(\ix{a},\exists,\ldots),$\\
         \>                       \>$\term(\ix{h},\forall,\ldots)))$
\end{tabbing}
\end{EXAMPLE}

There is another scoping option which instantiates $?S1$ to $[\ix{h}]$,
i.e. gives `every house' wide scope over both antecedent and ellipsis.
In this case the two terms, $\term(\ix{h}\ldots)$ in ellipsis and antecedent
are both discharged (i.e. bound) at the scope node $?S1$, rather than
being separately bound at the two copies of $?S2$
\begin{EXAMPLE}
\ENEW{flag-qlf2}
\begin{tabbing}
$[\ix{h}]$:and(\=$[\ix{c}]$:hang(\=$\term(\ix{c},\exists,\ldots),$\\
               \>                \>$\term(\ix{h},\forall,\ldots)),$\\
               \>$[\ix{a}]$:hang(\>$\term(\ix{a},\exists,\ldots),$\\
               \>                \>$\term(\ix{h},\forall,\ldots)))$
\end{tabbing}
\end{EXAMPLE}
(This has equivalent truth-conditions to \SREF{flag-qlf1}).\footnote{If
$\ix{c}$ is given wide scope over antecedent and ellipsis, the QLF is
rendered uninterpretable, which is as required.  As detailed in
section~\ref{semantics}, scoping \ix{c} discharges the term and its
index by substituting a variable for it.  But the ellipsis substitution
overrides this, substituting a new term and index, $\ix{a}$.  But there is
no way of discharging them.}

Besides illustrating scope parallelism, this is an example where DSP
have to resort to higher-order unification beyond second-order matching.
But no such increase in complexity is required under the present treatment.

\subsection{Strict and Sloppy Identity}

The notion of strict and sloppy identity is usually confined
to pronominal items occurring in antecedents and (implicitly) in
ellipses.\footnote{Also to pronouns of laziness.}
A standard example is
\begin{EXAMPLE}
\ENEW{ex21} John loves his mother, and Simon does too.
\end{EXAMPLE}
On the strict reading, Simon and John both love John's mother. The implicit
pronoun has been strictly identified with the pronoun in the antecedent to
pick out the same referent, John.  On the sloppy reading Simon loves Simon's
mother.
The implicit pronoun has been sloppily identified with its antecedent to
refer to something matching a similar description, i.e. the subject or agent
of the loving relation, Simon.

The sentence
\begin{EXAMPLE}
\ENEW{book} John read a book he owned, and so did Simon.
\end{EXAMPLE}
has three readings: John and Simon read the same book; John and Simon both
read a book belonging to John, though not necessarily the same one;
John reads one of John's books and Simon reads one of Simon's books.

Intuitively, the first reading arises from strictly identifying
the elliptical book with the antecedent book.  The second
arises from strictly identifying the pronouns, while
sloppily identifying the books.  The third from
sloppily identifying both the books and the pronouns. In the literature,
the first reading would not be viewed as a case of strict identity.
But this view emerges naturally from our treatment of substitutions, and
is arguably a more natural characterisation of the phenomena.

We need to distinguish between parallel and non-parallel
terms in ellipsis antecedents.  Parallel terms, like \lingex{John} in
the example above, are those that correspond terms appearing explicitly
in the ellipsis.  Non-parallel terms are those that do not have an explicit
parallel in the ellipsis. (Determining which terms are
parallel/non-parallel is touched on in section~\ref{parallelism}.)

For parallel terms, we have no choice about the ellipsis substitution.
We replace both the term and its index by the corresponding term
and index from the ellipsis.  But for {\em all} non-parallel terms
we have a choice between a strict or a sloppy substitution.\footnote{This
is true of the non-parallel $\term(\ix{h},\ldots)$ in example \SREF{flag-qlf};
but this added complication does not affect the basic account of
scope parallelism given earlier.}

A sloppy substitution involves substituting a new term index for the
old one.  This has the effect of reindexing the version of the term
occurring in the ellipsis, so that it refers to the same kind of thing
as the antecedent term but is not otherwise linked to it.

A strict substitution substitutes the term by its index.  In this
way, the version of the term occurring in the ellipsis is directly linked
to antecedent term.

To illustrate, an abbreviated QLF for the
antecedent \lingex{John read a book he owned} is
\begin{EXAMPLE}
\ENEW{book-ante}
\begin{tabbing}
{\it ?S}:\=\\
read(\\
   \>$\term(\ix{j}\ldots)$\\
   \>$\term($\=$\ix{b},\exists,$\\
   \>        \>$\lambda y.$\=$\mbox{book}(y)\wedge$\\
   \> \> \> $\mbox{own}(\term(\ix{h}\ldots
                              \reft{\ix{j}}),\;y),$\\
   \> \>$\ldots))$
\end{tabbing}
\end{EXAMPLE}
Here, we have left the scope node as an uninstantiated meta-variable
$?S$. The pronominal term $\ix{h}$ occurs in the restriction
of the book term $\ix{b}$.  The pronoun has been resolved to have
a contextual restriction, $\reft{\ix{j}}$, that co-indexes it with
the subject term.  Here, `$\reft{.}$' is a function that when applied
to an entity-denoting expression (e.g. a variable or constant) returns
the property of being identical to that entity; when it applies to a term
index, it returns an E-type property contextually linked to the term.

The ellipsis can be represented as
\begin{EXAMPLE}
\ENEW{book-ell}
$?P(\term(\ix{s},\;\exists,\;
\lambda y.\mbox{name}(y,\mbox{`Simon'}),\ldots))$
\end{EXAMPLE}
which is conjoined with the antecedent.

The three readings of \SREF{book} are illustrated below,
listing substitutions to be applied to the antecedent and cashing out
the results of their application, though omitting scope.

\begin{EXAMPLE}
\ENEW{be1}{\em Strict book}

$\{\ix{j}/\ix{s}, \; \term(\ix{j},\ldots)/\term(\ix{s},\ldots),$\\
$\; \term(\ix{b},\ldots)/\ix{b}, \; \ldots\}$

$\mbox{read}(\;\term(\;\ix{s}, \ldots), \; \ix{b})$
\end{EXAMPLE}
(a) Since all reference to the term $\ix{h}$ is removed by
the strict substitution on the term in which it occurs, it makes no
difference whether the pronoun is given a strict or a sloppy substitution.
(b) Strict substitution for the book leaves behind
an occurrence of the index $\ix{b}$ in the ellipsis.  For the QLF
to be interpretable, it is necessary to give the antecedent book term
wide scope over the ellipsis in order to discharge the index.

\begin{EXAMPLE}
\ENEW{be2}{\em Sloppy book, strict pronoun}

$\{\ix{j}/\ix{s}, \; \term(\ix{j},\ldots)/\term(\ix{s},\ldots),$\\
$\; \ix{b}/\ix{b1}, \; \term(\ix{h},\ldots)/\ix{h} \}$
\begin{tabbing}
re\=ad(\\
   \>$\term(\ix{s}\ldots)$\\
   \>$\term($\=$\ix{b1},\exists,$\\
   \>        \>$\lambda y.$\=$\mbox{book}(y)\wedge$\\
   \> \> \> $\mbox{own}(\ix{h},\; y)$\\
   \> \>$\ldots))$
\end{tabbing}
\end{EXAMPLE}
As above, the antecedent pronoun is constrained to be given wide
scope over the ellipsis, on pain of the index \ix{h} being undischargeable.
(Pronouns, like proper names,
are treated as contextually restricted quantifiers, where the contextual
restriction may limit the domain of quantification to one
individual.)

\begin{EXAMPLE}
\ENEW{be3}{\em Sloppy book, sloppy pronoun}

$\{\ix{j}/\ix{s}, \; \term(\ix{j},\ldots)/\term(\ix{s},\ldots),$\\
$\; \ix{b}/\ix{b1}, \; \ix{h}/\ix{h1} \}$

\begin{tabbing}
re\=ad(\\
   \>$\term(\ix{s}\ldots)$\\
   \>$\term($\=$\ix{b1},\exists,$\\
   \>        \>$\lambda y.$\=$\mbox{book}(y)\wedge$\\
   \> \> \> $\mbox{own}(\term(\ix{h1}\ldots
                              \reft{\ix{s}}),\;y),$\\
   \> \>$\ldots))$
\end{tabbing}
\end{EXAMPLE}
The index substitution from the primary term re-indexes the contextual
restriction of the pronoun.  It becomes coindexed with $\ix{s}$ instead of
$\ix{j}$.

DSP's account of the first reading of \SREF{book} is
significantly different from their account of the last two readings.
The first reading involves scoping the book quantifier before ellipsis
resolution.  The other two readings only scope the quantifier after
resolution, and differ in giving the pronoun a strict or a sloppy
interpretation. In our account
the choice of strict or sloppy substitutions for secondary
terms can constrain permissible quantifier scopings.\footnote{The converse
also holds.  Giving an antecedent term wide scope over the ellipsis
renders the choice of a strict or a sloppy substitution for it in the
ellipsis immaterial.  During semantic evaluation of the QLF, discharging
the antecedent through scoping will substitute out all occurrence of the
term and its index before ellipsis substitutions are applied. Note
though that this order dependence applies at the level of evaluating
QLFs, not constructing and resolving them.}
But the making of these
choices does not have to be interleaved in a precise order with the
scoping of quantifiers.

Moreover, the difference between strict and sloppy readings does not
depend on somehow being able to distinguish between primary and secondary
occurrences of terms with the same meaning. In DSP's representation of
the antecedent of \SREF{book}, both NPs `John' and `he' give rise to two
occurrences of the same term (a constant, $j$). The QLF representation
is able to distinguish between the primary and the secondary,
pronominal, reference to John.

\subsection{Other Phenomena}
Space precludes illustrating the substitutional approach through
further examples, though more are discussed in
\cite{AlsCarCro:clare,fracas:D9}.  The coverage is basically the same as
DSP's:

{\bf Antecedent Contained Deletion}: A sloppy substitution for
\lingex{every person that Simon did} in the sentence \lingex{John greeted
every person that Simon did} results in re-introducing the ellipsis in its
own resolution.  This leads to an uninterpretable cyclic QLF in much the same
way that DSP obtain a violation of the occurs check on sound unification.

{\bf Cascaded Ellipsis}: The number of readings obtained for
\lingex{John revised his paper before the teacher did, and then Simon did}
was used as a benchmark by DSP.  The approach here gets the four readings
identified by them as most plausible.  With slight modification, it gets
a fifth reading of marginal plausibility.  The modification is to allow
(strict) substitutions on terms not explicitly appearing in the ellipsis
antecedent --- i.e. the  implicit \lingex{his paper} in the second ellipsis
when resolving the third ellipsis.

We do not get a sixth, implausible reading, provided that in the first clause
\lingex{his} is resolved as being coindexed with the \term\ for \lingex{John};
i.e. that \lingex{John} and \lingex{his} do not both {\em independently}
refer to the same individual.  Kehler blocks this
reading in a similar manner. DSP block the reading by
a more artificial restriction on the depth of embedding of expressions
in logical forms; they lack the means for distinguishing between coindexed
and merely co-referential expressions.

{\bf Multiple VP Ellipsis}  Multiple VP ellipsis \cite{Gardent:auatmv}
poses problems at the level of determining which VP is the antecedent of
which ellipsis.  But at the level of incorporating elliptical material
once the antecedents have been determined, it appears to offer no
special problems.

{\bf Other Forms of Ellipsis}: Other forms of ellipsis, besides VP-ellipsis
can be handled substitutionally. For example,  NP-ellipsis
(e.g. \lingex{Who slept? John.}) is straightforwardly accommodated.
PP-ellipsis (e.g. \lingex{Who left on Tuesday? And on Wednesday?}) requires
substitutions for \form\ constructions in QLF (not described here)
representing prepositional phrases.

\subsection{Comparisons}

The use of terms and indices has parallels to proposals due
to Kehler and Kamp \cite{Kehler:adcafe,GawPet:aaqiss}. Kehler
adopts an analysis where (referential) arguments to verbs are represented
as related to a Davidsonian event via thematic role functions, e.g.
{\it agent(e)=john)}.  Pronouns typically refer to these functions,
e.g. {\it he=agent(e)}.  In VP ellipsis, strict identity corresponds
to copying the entire role assignment from the antecedent.  Sloppy identity
corresponds to copying the function, but applying it to the event of the
ellided clause.

For Kamp, strict identity involves copying the discourse
referent of the antecedent and identifying it with that of the
ellided pronoun.  Sloppy identity copies the conditions on the antecedent
discourse referent, and applies them to the discourse referent
of the ellided pronoun.

Neither Kamp nor Kehler extend their copying/substitution
mechanism to anything besides pronouns, as we have done.
In Kehler's case, it is hard to see how his role assignment functions
can be extended to deal with non-referential terms in the desired manner.
DRT's use of discourse referents to indicate scope suggests that Kamp's
treatment may be more readily extended in this manner; lists of discourse
referents at the top of DRS boxes are highly reminiscent of the index
lists in scope nodes.

\section{Semantic Evaluation}
\label{semantics}

Figure~\ref{fig} defines a valuation relation for the QLF fragment
used above, derived from \cite{AlsCro:msi,fracas:D8}.
\begin{figure}
{\small
Definition of ${\cal V}(\QLF,M,\varass,\substs,v)$
\begin{tabbing}
where \= $\QLF$ is a QLF expression\\
      \> $M$ is a model, $\langle O,F\rangle$\\
      \> $\varass$ is an assignment of values to variables\\
      \> $\substs$ is a set of substitutions\\
      \> $v$ is a value assigned to the QLF expression
\end{tabbing}
\begin{enumerate}

\item
Constant symbols, $c$: ${\cal V}(c,M,\varass,\substs,v)$ iff $F(c)=v$\\
(where $F$ is the interpretation function for non-logical constants
provided by $M$)
\label{qlf-const}

\item
Variables, $x$: ${\cal V}(x,M,\varass,\substs,v)$ iff $\varass(x)=v$
\label{qlf-var}

\item Reinterpretation:\\
${\cal V}(\QLF_1,M,\varass,\substs,v)$ iff ${\cal
V}(\QLF_2,M,\varass,\substs,v)$
where $\QLF_1 / \QLF_2 \in \substs$
\label{qlf-reinterp}

\item Merging reinterpretations:\\
${\cal V}(\QLF$$\mid$$\substs_1,M,\varass,\substs_2,v)$ if\\
${\cal V}(\QLF,M,\varass,\substs_1\Smerge\substs_2,v)$
\label{qlf-sub}

\item
Abstraction:\\
${\cal V}(\lambda x.\phi,M,\varass,\substs,h)$ if\\
$\phi \qlfsub \phi '$ and $h$ is such that
${\cal V}(\phi ',M,\varass^x_k,\substs,v)$ iff $h(k,v)$
\label{qlf-abs}

\item
Application:\\
${\cal V}(p(a_1,\ldots,a_n),M,\varass,\substs,P(A_1,\ldots,A_n))$ if\\
 $p(a_1,\ldots,a_n) \qlfsub p'(a'_1,\ldots,a'_n)$,\\
   ${\cal V}(p',M,\varass,\substs,P)$,\\
   ${\cal V}(a'_1,M,\varass,\substs,A_1)$,\\ $\ldots$, and\\
   ${\cal V}(a'_n,M,\varass,\substs,A_n)$
\label{qlf-app}

\item
$\qlfhat$-Application:\\
${\cal V}(X\qlfhat\phi\;(T), \;M,\varass,\substs,v)$ if\\
${\cal V}(\phi\mid\{X/T\}, \;M,\varass,\substs,v)$

\item
Scoped formula:\\
${\cal V}(Scope$:$\phi,M,\varass,\substs,v)$ if\\
${\cal V}(Q'(R',\phi '),M,\varass,\substs,v)$\\
where\\
a)$\phi$ is a formula containing the term, $T_0$,
${\tt term(}I_0,Q_0,R_0,P_0{\tt )}$\\
b) $\Replace(T_0,\substs) = T = {\tt term(}I,C,R,Q,P{\tt )}$ \\
c) $Scope \qlfsub [I,\ldots]$\\
d) $R'$ is $\lambda x.(\verb!and(!R(x),P'(x))\mid\{I/x\})$\\
e) $\phi '$ is $\lambda x.([\ldots]\mbox{:}\phi\mid\{T/x,I/x\})$
\label{qlf-scope}

\end{enumerate}

Operations on substitutions:
\begin{itemize}
\item
$\substs_1\Smerge\substs_2$ combines two sets of substitutions.
This is like set union, except that where $\substs_1$ and $\substs_2$
both substitute for a particular item, the substitution from $\substs_1$
is retained and not that in $\substs_2$.
\item
$\Replace(Old,\substs)$ returns $New$ if $Old/New \in \substs$ and
otherwise $Old$.
\end{itemize}} 
\caption{QLF Evaluation Rules}
\label{fig}
\end{figure}
If a QLF expression contains uninstantiated meta-variables, the valuation
relation can associate more than one value with the expression.  In the
case of formulas, they may be given both the values true and false,
corresponding to the formula being true under one possible resolution
and false under another.  A subsumption ordering over QLFS, $\qlfsub$,
is employed in the evaluation rules, in effect to propose possible
instantiations for meta-variables (the rule fragment only allows for
scope meta-variables, but \cite{fracas:D8} describes the more general case
where other kinds of meta-variable are permitted).  A partially instantiated
QLF therefore effectively specifies a set of possible evaluations (or
semantic compositions).  As the QLF becomes more instantiated, the set of
possible evaluations narrows towards a singleton.

It is also possible for a QLF to be uninterpretable; to specify no possible
evaluation.  Thus, no rules are given for evaluating terms or their indices
in isolation.  They must first be discharged by the scoping rule, which
substitutes the terms and indices by $\lambda$-bound variables.  Inappropriate
scoping may leave undischarged and hence uninterpretable terms and indices
(which accounts for the so-called free-variable
and vacuous quantification constraints on scope \cite{AlsCro:msi}).

The non-deterministic nature of evaluation and the role of substitutions
draws us to conclude that ellipsis substitutions operate on (descriptions of)
the semantic compositions, not the results of such compositions.

\section{Parallelism and Inference}
\label{parallelism}

Selecting ellipsis antecedents and parallel elements within them is an
open problem
\cite{Prust:adsvaa,PruSchBer:dgavpa,Kehler:teoeci,GroBreMan:puagid}.
Our approach to parallelism is perhaps heavy-handed, but in the absence of
a clear solutions, possibly more flexible. The QLFs shown
above omitted category information present in \terms\ and
\forms.\footnote{ \forms\ are described in \cite{AlsCro:msi}.}
Categories are sets of feature value equations
containing syntactic information relevant to determining how
uninstantiated meta-variables can be resolved.

Tense in VP-ellipsis illustrates how categories can be put to work. In
\begin{EXAMPLE}
\ENEW{tense} I enjoyed it. And so will you
\end{EXAMPLE}
the ellipsis is contained within a form expression whose category is
\begin{quoteqlf}\begin{verbatim}
vp_ellipsis[tense=inf,modal=will,perfect=_,
            progressive=_,pol=pos,...]
\end{verbatim}\end{quoteqlf}
This states the syntactic tense, aspect and polarity marked on the ellipsis
(underscores indicate lack of specification).
The category constrains resolution to look for verb phrase/sentence sources,
which come wrapped in \forms\ with categories like
\begin{quoteqlf}\begin{verbatim}
vp[tense=past,modal=no,perfect=no,
   progressive=no,pol=pos,...]
\end{verbatim}\end{quoteqlf}
Heuristics similar to those described by Hardt \shortcite{Hardt:aafve}
may be used for this.
The category also says that, for this kind of VP
match\footnote{Not all VP ellipses have VP antecedents.}, the term in
the antecedent whose category identifies it
as being the subject should be treated as parallel to the explicit term
in the ellipsis.

As this example illustrates, tense and aspect on ellipsis and antecedent do
not have to agree.  When this is so, the antecedent and ellipsis
categories are both used to determine what \form\ should be
substituted for the antecedent \form. This comprises the restriction of
the antecedent \form\ and a new category constructed by taking the
features of the antecedent category, unless overridden by those on the
ellipsis---a kind of (monotonic) priority union \cite{GroBreMan:puagid}
except using skeptical as opposed to credulous default unification
\cite{Carpenter:sacduw}.  When a new category
is constructed for the antecedent, any tense resolutions also need to be
undone, since the original ones may no longer be appropriate for the
revised category.  One thus merges the category information from source and
antecedent to determine what verb phrase form should be substituted for the
original.  In this case, it will have a category
\begin{quoteqlf}\begin{verbatim}
vp[tense=inf,modal=will,perfect=no,
   progressive=no,pol=neg,...]
\end{verbatim}\end{quoteqlf}

A more general question is whether all ellipses involve recompositions,
with variants, of linguistic antecedents.  There are cases where
a degree of inference seems to be required:
\begin{EXAMPLE}
\ENEW{inf}
We spent six weeks living in France, eating French food and speaking French,
as we did in Austria the year before.
\end{EXAMPLE}
(one must apply the knowledge that Austrians speak German to correctly
interpret the ellipsis).  Pulman's \shortcite{Pulman:actcd} equational
treatment of context-dependency suggests one method of dealing with such cases.
But it remains to be seen how readily the equations used for ellipsis
here can be integrated into Pulman's framework.

\section{Conclusions: Interpretation as Description}
\label{conc}

The substitutional treatment of ellipsis presented here has broadly
the same coverage as DSP's higher-order unification treatment,
but has the computational advantages of (i) not requiring order-sensitive
interleaving of different resolution operations, and (ii) not requiring
greater than second-order matching for dealing with quantifiers.
In addition, it cures a slight overgeneration problem in DSP's account.

It has been claimed that these advantages arise from viewing semantic
interpretation as a process of building descriptions of semantic
compositions.  To conclude, a few
further arguments for this view, that are independent of any particular
proposals for dealing with ellipsis.

{\bf Order-Independence:} One of the reasons for the computational
success of unification-based syntactic formalisms is the order-independence
of parser/generator operations they permit.  If one looks at the
order-sensitive nature of the operations of semantic compositions, they
provide a poor starting point for a treatment of semantics enjoying
similar computational success.  But semantic interpretation, viewed as
building a description of the intended composition, is a better prospect.

{\bf Context-Sensitivity:}  The truth values of many (all?) sentences
undeniably depend on context.  Context-dependence may enter either
at the interpretive mapping from sentence to meaning and/or the evaluative
mapping from meaning (and the world) to truth-values.
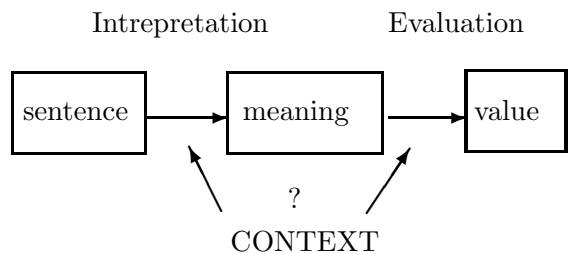
\begin{quoteqlf}
\setlength{\unitlength}{0.0125in}%
{\tiny
\begin{picture}(231,104)(20,709)
\thicklines
\put( 75,765){\vector( 1, 0){ 35}}
\put( 20,750){\framebox(55,33){}}
\put(107,726){\vector(-1, 2){ 13.400}}
\put(168,725){\vector( 2, 3){ 18}}
\put(177,765){\vector( 1, 0){ 33}}
\put(210,751){\framebox(41,33){}}
\put(110,750){\framebox(64,33){}}
\put( 53,801){\makebox(0,0)[lb]{\raisebox{0pt}[0pt][0pt]{\elvrm
Intrepretation}}}
\put(177,801){\makebox(0,0)[lb]{\raisebox{0pt}[0pt][0pt]{\elvrm Evaluation}}}
\put(111,709){\makebox(0,0)[lb]{\raisebox{0pt}[0pt][0pt]{\elvrm CONTEXT}}}
\put( 24,764){\makebox(0,0)[lb]{\raisebox{0pt}[0pt][0pt]{\elvrm sentence}}}
\put(135,727){\makebox(0,0)[lb]{\raisebox{0pt}[0pt][0pt]{\elvrm ?}}}
\put(213,764){\makebox(0,0)[lb]{\raisebox{0pt}[0pt][0pt]{\elvrm value}}}
\put(116,764){\makebox(0,0)[lb]{\raisebox{0pt}[0pt][0pt]{\elvrm meaning}}}
\end{picture}}
\end{quoteqlf}
The more that context-dependence enters into the interpretive mapping
(so that meanings are correspondingly more context-independent), the
harder it is to maintain a principle of strict compositionality
in interpretation. The syntactic structure underspecifies the intended
composition, so that the meanings of some constituents (e.g. pronouns)
and the mode of combination of other (e.g. quantifiers) are not fully
specified. Further contextual information is required to fill the gaps.
Again, interpretation seen as description building sits easily with this.

{\bf Preserving Information:} Focusing exclusively on the results of
semantic composition, i.e. meanings, can ignore differences in how those
meanings were derived that can be linguistically significant (e.g.
co-referential vs co-indexed terms).  If this information is not to be lost,
some way of referring to the structure of the compositions, as well as to
their results, seems to be required.

\section*{Acknowledgements}
The initial implementation of the work described here was carried out
as part of the CLARE project, DTI IED4/1/1165.  The writing of this paper
was supported in part by the FraCaS project, LRE 62-051.  I would especially
like to thank Hiyan Alshawi and Steve Pulman for help and advice on
topics relating to this paper.  I have also benefited from conversations
with Claire Grover, Ian Lewin and Massimo Poesio.


\small
\begin{thebibliography}{}

\bibitem[\protect\citename{Alshawi and Crouch}1992]{AlsCro:msi}
Hiyan Alshawi and Richard Crouch.
\newblock 1992.
\newblock Monotonic semantic interpretation.
\newblock In {\em Proceedings 30th Annual Meeting of the Association for
  Computational Linguistics}, pages 32--38.

\bibitem[\protect\citename{Alshawi \bgroup et al.\egroup
  }1992]{AlsCarCro:clare}
Hiyan Alshawi, David Carter, Richard Crouch, Stephen Pulman, Manny Rayner, and
  Arnold Smith.
\newblock 1992.
\newblock Clare: A contextual reasoning and cooperative response framework for
  the core language engine.
\newblock Technical Report CRC-028, SRI International, Cambridge Computer
  Science Research Centre.
\newblock Available by anonymous ftp from ftp.ai.sri.com,
  pub/sri-cam/reports/crc028.ps.Z; also cmp-lg.

\bibitem[\protect\citename{Carpenter}1993]{Carpenter:sacduw}
Bob Carpenter.
\newblock 1993.
\newblock Skeptical and credulous default unification with applications to
  templates and inheritance.
\newblock In E.~Briscoe, V.~de~Paiva, and A.~Copestake, editors, {\em
  Inheritance, Defaults and the Lexicon}, pages 13--37. Cambridge University
  Press.

\bibitem[\protect\citename{Cooper \bgroup et al.\egroup }1994a]{fracas:D8}
Robin Cooper, Richard Crouch, Jan van Eijck, Chris Fox, Josef van Genabith, Jan
  Jaspars, Hans Kamp, Manfred Pinkal, Massimo Poesio, and Stephen Pulman.
\newblock 1994a.
\newblock Describing the approaches.
\newblock FraCaS deliverable, D8. Available by anonymous ftp from
  ftp.cogsci.ed.ac.uk, pub/FRACAS/del8.ps.gz.

\bibitem[\protect\citename{Cooper \bgroup et al.\egroup }1994b]{fracas:D9}
Robin Cooper et al.
\newblock 1994b.
\newblock Evaluating the descriptive capabilities of semantic theories.
\newblock FraCaS deliverable, D9. Available by anonymous ftp from
  ftp.cogsci.ed.ac.uk, pub/FRACAS/del9.ps.gz.


\bibitem[\protect\citename{Dalrymple \bgroup et al.\egroup
  }1991]{DalShiPer:eahu}
Mary Dalrymple, Stuart~M. Shieber, and Fernando C.~N. Pereira.
\newblock 1991.
\newblock Ellipsis and higher-order unification.
\newblock {\em Linguistics and Philosophy}, 14:399--452.

\bibitem[\protect\citename{Gardent}1993]{Gardent:auatmv}
Claire Gardent.
\newblock 1993.
\newblock A unification-based approach to multiple vp ellipsis resolution.
\newblock In {\em Proceedings 6th European ACL}, pages 139--148.

\bibitem[\protect\citename{Gawron and Peters}1990]{GawPet:aaqiss}
Mark Gawron and Stanley Peters.
\newblock 1990.
\newblock {\em Anaphora and Quantification in Situation Semantics}.
\newblock Number~19 in CSLI Lecture Notes. CSLI, Stanford, CA.

\bibitem[\protect\citename{Grover \bgroup et al.\egroup
  }1994]{GroBreMan:puagid}
Claire Grover, Chris Brew, Suresh Manandhar, and Marc Moens.
\newblock 1994.
\newblock Priority union and generalization in discourse grammars.
\newblock In {\em Proceedings 32nd Annual Meeting of the Association for
  Computational Linguistics}, pages 17--24.

\bibitem[\protect\citename{Hardt}1992]{Hardt:aafve}
Daniel Hardt.
\newblock 1992.
\newblock An algorithm for vp ellipsis.
\newblock In {\em Proceedings 30th Annual Meeting of the Association for
  Computational Linguistics}, pages 9--14.

\bibitem[\protect\citename{Kehler}1993a]{Kehler:adcafe}
Andrew Kehler.
\newblock 1993a.
\newblock A discourse copying algorithm for ellipsis and anaphora resolution.
\newblock In {\em Proceedings 6th European ACL}, pages 203--212.

\bibitem[\protect\citename{Kehler}1993b]{Kehler:teoeci}
Andrew Kehler.
\newblock 1993b.
\newblock The effect of establishing coherence in ellipsis and anaphora
  resolution.
\newblock In {\em Proceedings 31st Annual Meeting of the Association for
  Computational Linguistics}, pages 62--69.

\bibitem[\protect\citename{Nerbonne}1991]{Nerbonne:cs}
John Nerbonne.
\newblock 1991.
\newblock Constraint-based semantics.
\newblock In {\em Proceedings 8th Amsterdam Colloquium}, pages 425--443.


\bibitem[\protect\citename{Pr\"{u}st \bgroup et al.\egroup
  }1994]{PruSchBer:dgavpa}
Hub Pr\"{u}st, Remko Scha, and Martin van~den Berg.
\newblock 1994.
\newblock Discourse grammar and verb phrase anaphora.
\newblock {\em Linguistics and Philosophy}, 17:261--327.

\bibitem[\protect\citename{Pr{\"u}st}1992]{Prust:adsvaa}
Hub Pr{\"u}st.
\newblock 1992.
\newblock {\em On Discourse Structure, VP Anaphora and Gapping}.
\newblock {Ph.D.} thesis, University of Amsterdam.

\bibitem[\protect\citename{Pulman}1994]{Pulman:actcd}
Stephen Pulman.
\newblock 1994.
\newblock A computational theory of context dependency.
\newblock In {\em Proceedings of the International Workshop on Computational
  Semantics}, pages 161--171, Tilburg.

\end{thebibliography}
\end{document}